\documentclass[journal=jacsat,manuscript=article]{achemso}
\usepackage[T1]{fontenc}
\usepackage{array}
\newcolumntype{L}[1]{>{\raggedright\arraybackslash}p{#1}}

\usepackage{hyperref}
\usepackage{tabularx}
\usepackage{array}
\usepackage{graphicx}
\usepackage{subcaption}
\usepackage{float}
\usepackage{stfloats}
\usepackage{adjustbox}
\usepackage{siunitx}
\usepackage{textcomp}
\usepackage{bbm}
\usepackage{amsfonts}
\usepackage{mciteplus}
\mciteErrorOnUnknownfalse
\usepackage{listings}
\usepackage{xcolor}
\usepackage{longtable}
\usepackage{multirow}
\usepackage{titletoc}
\definecolor{jsonbg}{gray}{0.97}
\lstdefinestyle{json}{
  basicstyle=\ttfamily\scriptsize,
  backgroundcolor=\color{jsonbg},
  frame=single,
  rulecolor=\color{gray!60},
  breaklines=true,
  showstringspaces=false,
  columns=fullflexible,
  keepspaces=true,
  tabsize=2,
  xleftmargin=1em,
  xrightmargin=1em,
  aboveskip=0.8em,
  belowskip=0.8em,
}

\definecolor{promptBlueBg}{RGB}{234,244,252}
\definecolor{promptBlueFr}{RGB}{31,97,141}
\definecolor{promptOrangeBg}{RGB}{253,244,228}
\definecolor{promptOrangeFr}{RGB}{194,108,32}
\definecolor{promptPurpleBg}{RGB}{244,238,252}
\definecolor{promptPurpleFr}{RGB}{110,62,153}
\definecolor{promptGreenBg}{RGB}{236,247,238}
\definecolor{promptGreenFr}{RGB}{39,120,72}

\lstdefinestyle{promptBlue}{
  basicstyle=\ttfamily\footnotesize,
  backgroundcolor=\color{promptBlueBg},
  frame=single,
  rulecolor=\color{promptBlueFr},
  framesep=6pt,
  breaklines=true,
  breakatwhitespace=false,
  columns=flexible,
  keepspaces=true,
  showstringspaces=false,
  upquote=true,
  numbers=none,
  literate={~}{{\textasciitilde}}1
}
\lstdefinestyle{promptOrange}{
  basicstyle=\ttfamily\footnotesize,
  backgroundcolor=\color{promptOrangeBg},
  frame=single,
  rulecolor=\color{promptOrangeFr},
  framesep=6pt,
  breaklines=true,
  breakatwhitespace=false,
  columns=flexible,
  keepspaces=true,
  showstringspaces=false,
  upquote=true,
  numbers=none,
  literate={~}{{\textasciitilde}}1
}
\lstdefinestyle{promptPurple}{
  basicstyle=\ttfamily\footnotesize,
  backgroundcolor=\color{promptPurpleBg},
  frame=single,
  rulecolor=\color{promptPurpleFr},
  framesep=6pt,
  breaklines=true,
  breakatwhitespace=false,
  columns=flexible,
  keepspaces=true,
  showstringspaces=false,
  upquote=true,
  numbers=none,
  literate={~}{{\textasciitilde}}1
}
\lstdefinestyle{promptGreen}{
  basicstyle=\ttfamily\footnotesize,
  backgroundcolor=\color{promptGreenBg},
  frame=single,
  rulecolor=\color{promptGreenFr},
  framesep=6pt,
  breaklines=true,
  breakatwhitespace=false,
  columns=flexible,
  keepspaces=true,
  showstringspaces=false,
  upquote=true,
  numbers=none,
  literate={~}{{\textasciitilde}}1
}

\newcommand{\includegraphicsorplaceholder}[2][]{%
  \IfFileExists{#2}{%
    \includegraphics[#1]{#2}%
  }{%
    \fbox{%
      \parbox[c][0.26\textheight][c]{0.94\linewidth}{%
        \centering\small Figure file unavailable at compile time:\\
        \path{#2}%
      }%
    }%
  }%
}

\SectionNumbersOn

\usepackage[version=3]{mhchem} 
\usepackage{booktabs}
\usepackage{pifont} 

\author{Achuth Chandrasekhar}
\affiliation[CMU-MECHE]
{Mechanical Engineering, Carnegie Mellon University, Pittsburgh, PA 15213, USA}

\author{Omid Barati Farimani}
\affiliation[CMU-MECHE]
{Mechanical Engineering, Carnegie Mellon University, Pittsburgh, PA 15213, USA}
\author{Radheesh Sharma Meda}
\author{Amir Barati Farimani}
\email{barati@cmu.edu}
\affiliation[CMU-MECHE]
{Mechanical Engineering, Carnegie Mellon University, Pittsburgh, PA 15213, USA}
\alsoaffiliation[CMU-BME]
{Biomedical Engineering, Carnegie Mellon University, Pittsburgh, PA 15213, USA}
\alsoaffiliation[CMU-CHE]
{Chemical Engineering, Carnegie Mellon University, Pittsburgh, PA 15213, USA}
\alsoaffiliation[CMU-MSE]
{Materials Science and Engineering, Carnegie Mellon University, Pittsburgh, PA 15213, USA}
\alsoaffiliation[CMU-ML]
{Machine Learning Department, Carnegie Mellon University, Pittsburgh, PA 15213, USA}

\title[An \textsf{achemso} demo]
  {Material Database Agent: A Multimodal Agentic Framework for Scientific Literature Mining}

\DeclareUnicodeCharacter{202F}{\,}
\begin{document}


\begin{abstract}
Materials science workflows rely on structured and unstructured data from the vast body of available scientific literature. However, most of the experimental details remain buried in text, tables, graphs and figures. Thus, constructing databases that incorporate this data is a manual, time-consuming, and hard-to-scale process. Multimodal large language models have made it feasible to extract information from text and scientific figures with high speed and accuracy. This opens the possibility of an AI system that can create production-scale material databases. Material Database Agent (MDA) is a modular, multi-agent system architecture for converting research literature into structured databases. MDA accepts article PDFs as input, which are subsequently processed in parallel into markdown files and figures. Multiple sub-agents read these markdown files and figures in parallel to assemble sub-databases for each paper. These sub-databases are then compiled into a single tabular database by an agent. As opposed to using either
a rule-based approach or a single-pass pipeline for extracting information, MDA is a specialized architecture for transforming the literature into a database in the field
of materials science. More generally, this study provides a basis for positioning multimodal agentic information extraction as a viable means for constructing next-generation scientific databases from the primary literature.

\end{abstract}

\section*{Introduction}
\label{sec:introduction}
Experimental and computational research in materials science has given rise to a vast body of knowledge\cite{meidani2021titanium,barati2024fast,cao2020single}. Unfortunately, the majority of information that exists nowadays is geared towards facilitating comprehension\cite{ikegwu2022big,quinones2025data}. The ability to assemble databases, evaluate comparative studies, and develop data-driven models is limited by access to human labor and physical time constraints. Consequently, the need for automated literature extraction is both technically motivated and increasingly necessary for the scalability of materials discovery, benchmarking, and validation\cite{cao2024machine}. 

Early materials information extraction depended on basic NLP methods, including named entity recognition, regular expressions, rule-based parsers and pre-LLM neural network architectures. These approaches demonstrated the feasibility of large-scale scientific text mining and supported the development of valuable domain-specific databases \cite{weston2019named,olivetti2020data,zhu2022pdfdataextractor, clark2016pdffigures, tkaczyk2015cermine, luan2018multi, gupta2022matscibert,wilary2023reactiondataextractor}.


ChemDataExtractor by Sierepeklis et al.\cite{sierepeklis2022thermoelectric} applies multiple pre-LLM NLP methods for textual data extraction, but these approaches remain limited. They rely on extensive schema design, hand-crafted rules, and task-specific tuning, and often struggle with varied phrasing, non-standard formatting, and multimodal evidence across text and figures. As a result, the field can extract specific data types effectively, but still lacks flexible general-purpose systems that are robust to document structure and reporting style.

Recent advances in large language models have reshaped this landscape \cite{chandrasekhar2024amgpt,chandrasekhar2025nanogpt,cao2024machine, sayeed2025knowmat, turan2026revolutionizing, polak2025leveraging, stewart2026graphagents, jadhav2025llm, badrinarayanan2025mofgpt, ock2023catalyst}. Conversational extraction workflows, such as those of Polak and Morgan (2024), show that LLMs can recover materials data accurately while reducing the development burden of traditional extraction pipelines \cite{polak2024extracting}.
nanoMINER by Odobesku et al. presents a multi-agent framework for multimodal data extraction using GPT-4o. This is a pathbreaking work that shows the high data-extraction accuracy of multi-agent systems. However, GPT-4o exhibits below-human levels of performance in multimodal understanding, as evidenced by its difficulties with complex plots \cite{odobesku2025agent}. Li et al. introduce the SciEx framework, which extracts data from text, tables and figures in a structured manner, but still falls short of producing reliable material databases for production-level applications as acknowledged by the authors \cite{li2025exploring}. Ghosh et al. only extract textual data from XML and HTML files, using older LLMs such as GPT-4.1 and Gemini-1.5-Pro, and acknowledge that accuracy is low for material properties that are highly represented in plots and figures \cite{ghosh2026llm}. Rameshbabu et al.\cite{rameshbabu2026papers} extract material properties from text and figures, but figures are considered low-priority sources and only two models (Gemini 3.0 and Opus 4.5) are evaluated in this paper.

Collectively, these studies represent a new
paradigm: literature extraction is now plausibly multimodal, prompt-driven, and significantly more adaptable than prior extraction systems. Timelines for extraction have also shrunk by orders of magnitude (from many months of manual human labor to mere hours). While many studies focus on aspects of the extraction problem (e.g. extracting sentences from text, digitizing plots, identifying entities
for a given property class, etc.), few studies address the fundamental question of how to systematically decompose a full research paper, route its components for parallel processing, control its quality, and reconstruct them into a single database entry format \cite{adhikary2024case, rombach2025deep}. Furthermore,
this gap is particularly pronounced in materials papers, where the pertinent evidence
is unevenly distributed across sections, captions, embedded images, and supplemental-style fragments. Moreover, a single document may contain multiple materials, multiple processing
conditions, and/or multiple target properties. Placing the entirety of a paper into a single model context is often inefficient, brittle, or slow, and when document parsing itself becomes a computational bottleneck, this inefficiency is compounded \cite{bai2024beyond,kambhampati2024position,bansal2025let}. 

Concurrently, Chong and Colindres \cite{Chong2026-xy} introduce LitXBench, a benchmarking framework that advances evaluation rigor for experiment-level extraction from alloy papers through code-based ground-truth representations. However, LitXBench considers only textual information and does not extract data from figures, a limitation the authors acknowledge leads to ground-truth inaccuracies when key measurements are embedded in plots rather than text. This constraint is shared to varying degrees by all existing extraction frameworks and reinforces the need for architectures that are natively multimodal.

In this work, we address this gap with MDA, an agentic workflow\cite{nigam2026polymer, pak2026agentic, ock2026adsorb, han2025tdflow, ock2025large, chandrasekhar2507automating, zhao2026polyjarvis, chaudhari2025modular, zeng2025llm, jadhav2026large}designed to support end-to-end conversion of research articles into structured materials databases. The workflow commences with PDF ingestion and extraction via a primary agent, after which the parsed article is decomposed into localized subdirectories that
preserve the relevant markdown and image context. Each decomposed unit is subsequently processed in parallel by a doc-writer sub-agent that extracts material parameters into structured JSON outputs. Finally, a csv-writer agent aggregates the intermediate outputs into a unified
database. The rationale for this design is straightforward. Accurate scientific extraction at
scale requires not only effective models, but also a document architecture that minimizes
context interference, preserves multimodal locality, and facilitates modular aggregation. By decomposing a paper into smaller units of evidence-bearing content before extraction, MDA aims to improve robustness, throughput, and adaptability to new schemas and new materials domains. The architecture is domain-agnostic and scales across material systems, data formats, and backbone models.

We test this process on two different datasets: 
\begin{enumerate}
    \item MeltpoolNet's experimental dataset for powder bed fusion additive manufacturing\cite{akbari2022meltpoolnet}.
    \item A database of high-entropy alloys and complex concentrated alloys\cite{rahaman2018pressure}.
\end{enumerate}

Accordingly, our contribution is not merely another extraction tool, but an agentic architecture for literature mining that leverages state-of-the-art multimodal models with near-human document understanding.

\section*{Results and Discussion}

\begin{figure}[hbt!]

\includegraphics[width=1.0\linewidth]{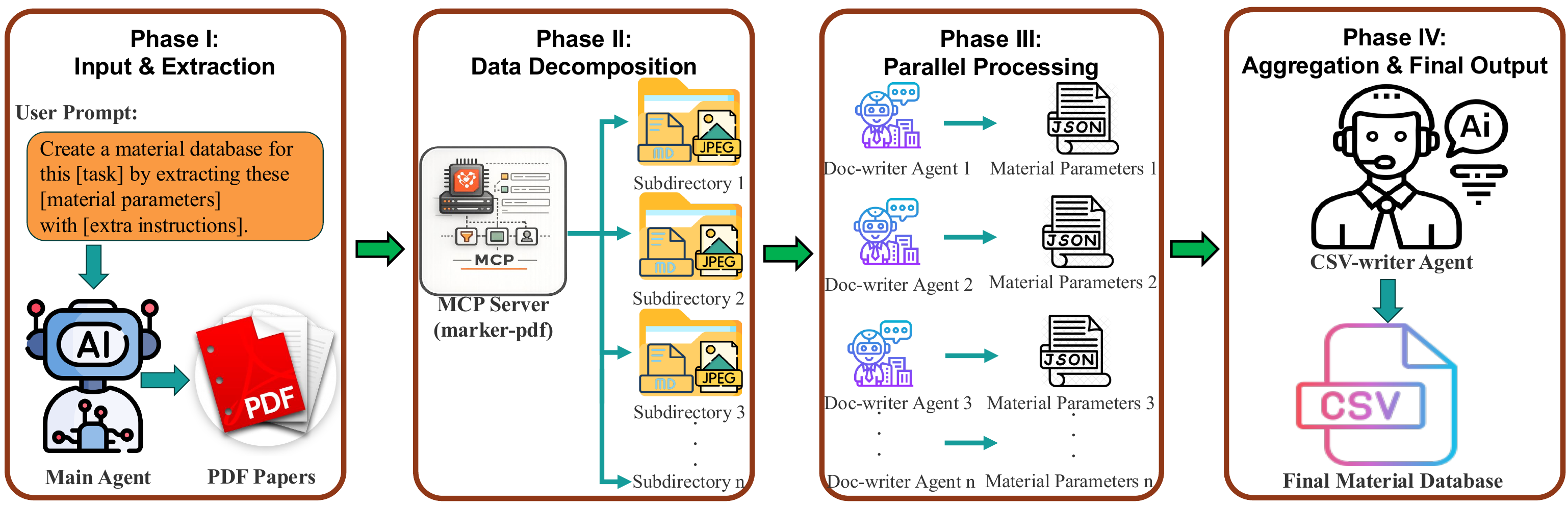}
\caption{Material Database Agent workflow. Overview of the four-stage pipeline: PDF ingestion, document decomposition into structured text and images, parallel sub-agent processing, and aggregation into a unified CSV material database.}
\label{fig:react}
\end{figure}

\subsection*{Agent operation workflow}
MDA's workflow is shown in Figure \ref{fig:react} and consists of four sequential modules. These are for input and extraction, document decomposition, parallel information processing and database aggregation. The workflow begins with the acquisition of relevant PDFs by the user. After collection, the location of the PDFs is given to the main agent that acts as a centralized supervisor for the entire system.

The main agent feeds the PDFs into the MCP server powered by the open-source marker-pdf tool\cite{paruchuri2024marker}. This tool parses each PDF with specialized document OCR and segmentation software. This converts them into markdown files for textual content and JPEG images for the various figures. This step is important because relevant data for the final database is frequently interspersed among the text, plots and graphs, rather than being contained in one uniform format. Also, tabular data has specific structures that need to be preserved for accurate data collection, and this is achieved by the use of the markdown format.

Once parsing and document deconstruction are completed, the output files are organized into subdirectories. Each subdirectory corresponds to one particular source PDF and contains the markdown text and images. 

In the third phase of this workflow, multiple parallel doc-writer sub-agents with their own independent context windows read the contents of each subdirectory. The spawning of these sub-agents is left to the discretion of the main agent. These sub-agents extract material properties and other relevant information to create text files in each subdirectory. These text files contain the extracted information in a predefined JSON format. This kind of "agent parallelism" allows for the scalable extraction of information in a multimodal setting, without decreasing accuracy or overflowing the available context length.

In the fourth and final phase of this workflow, all of the intermediate JSON-formatted text files are read by a single csv-writer sub-agent which compiles and consolidates the information snippets into a single, final material database csv file (Figure \ref{fig:react}). The descriptions of these various agents employed are given in Section S2 and the system prompts to direct their proper functioning are detailed in Section S3.

In the aggregation phase, the distributed outputs of the doc-writer sub-agents are consolidated into a unified tabular resource, enabling compatibility with downstream analysis and benchmarking.
The human-in-the-loop aspect of the architecture, illustrated in Figure \ref{fig:debye-extraction}, shows how the user provides prompts to the main agent, which then coordinates the MCP server and specialized sub-agents. This highlights that MDA is not purely an automated pipeline, but a prompt-driven and steerable multi-agent system whose behavior can be guided by the user.

\subsection*{Comparison of multimodal chart extraction across model families}

\begin{figure}[hbt!]
\centering
\includegraphics[width=1\linewidth]{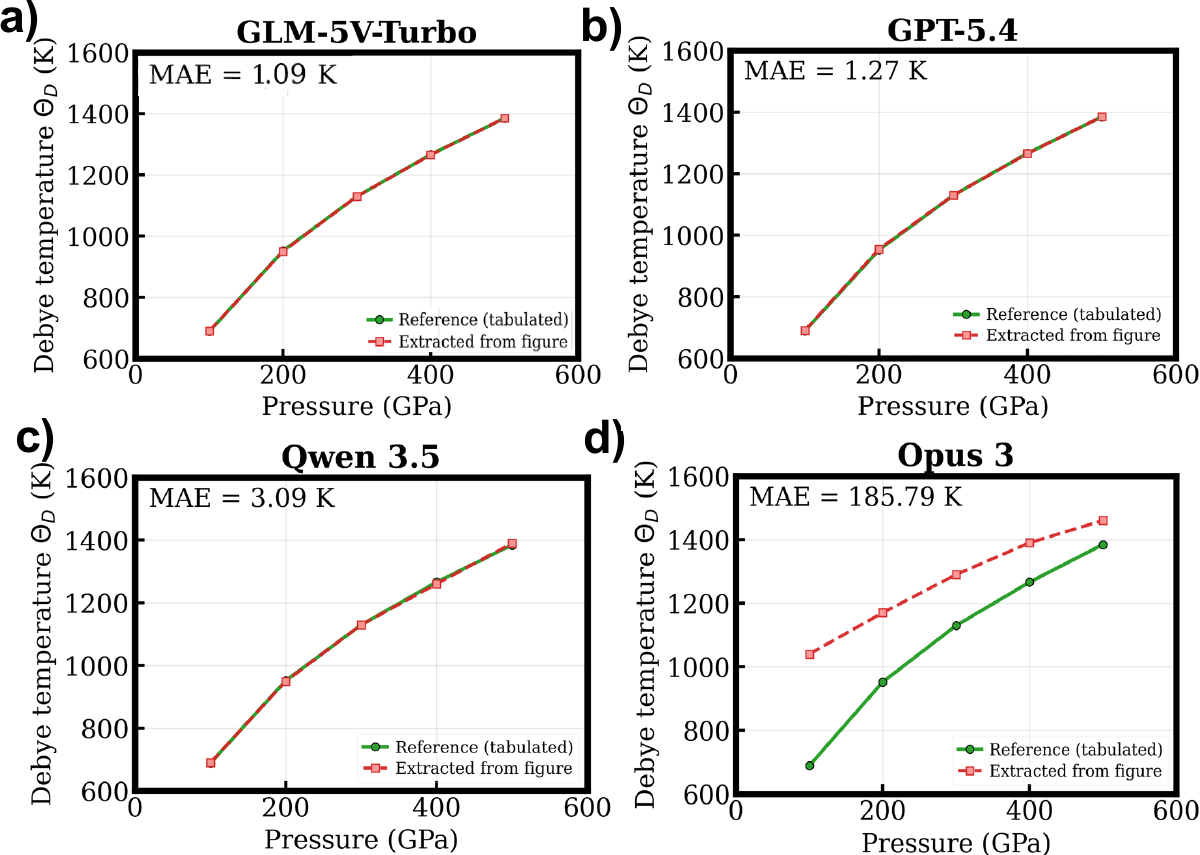}
\caption{Extracted versus tabulated Debye temperature for cubic \cite{rahaman2018pressure}, Table~4 and Figure~4a reference). Subpanels \textbf{(a)}--\textbf{(d)} use identical pressure and Debye-temperature axis limits. Ordering follows increasing MAE on this five-point task from \textbf{(a)} (lowest error) to \textbf{(d)} (largest error).}
\label{fig:debye-extraction}
\end{figure}

We evaluated and compared the multimodal extraction capabilities of various LLMs for a single materials science-based plot: the Debye temperature vs pressure curve for cubic \ce{Na2He} using the data from Table 4 of Rahaman et al., ``Pressure-dependent mechanical and thermodynamic properties of newly discovered cubic \ce{Na2He}''\cite{rahaman2018pressure}, which was used to plot Figure 4a of the same paper. The visual task was defined as the recovery of the exact points from the published plot. For each LLM, we report the mean absolute error (MAE) between the tabulated reference and the model-extracted values of the Debye temperature, at five pressure values ranging from 100 to 500 GPa with a step size of 100 GPa. Figure~\ref{fig:debye-extraction} shows the comparison plots for original and extracted data for each backbone model. Extra plots can be found in Section S4.

The models used here were released from 2024 to 2026 and span different generations. Claude Opus 3 was released in March 2024. Claude Sonnet 4 was made public in May 2025. OpenAI GPT-5.2 and GPT-5.4 were released in December 2025 and March 2026 respectively. Claude Opus 4.6 was released in February 2026 in the same time slot as the latter two. In this work, we use the MMMU (Massive Multi-discipline Multimodal Understanding and Reasoning Benchmark), a benchmark by Yue et al.\cite{yue2024mmmu}, to evaluate multimodal understanding and reasoning on challenging academic tasks. Table~\ref{tab:debye-extraction-mmmu} shows the MMMU (val/pro) scores and the extraction MAE for the various models tested. MMMU-pro was released in September 2024, so scores for the older models on that subset are not available. However, the scores available are still representative of overall multimodal performance.

On this task, the latest LLMs such as GPT-5.2, Opus 4.6, GPT-5.4 and GLM 5V Turbo achieve much smaller MAEs (3.44, 1.61, 1.27, 1.09 respectively), and their extracted points are able to reproduce the original plot with very high accuracy. These LLMs also show very high scores on the MMMU benchmark. An earlier, more primitive LLM like Opus 3 records a catastrophic MAE of $\approx$~186~K. This shows that a score on the MMMU (val/pro) benchmark is highly correlated with performance levels on plot digitization and data extraction. This case study demonstrates the ability of the latest LLMs to reconstruct literature data from plots, which is more of a bottleneck compared to simple textual extraction.

\begin{table}[htbp]
\centering
\caption{Multimodal Debye-temperature chart extraction for cubic \ce{Na2He} (Rahaman et al.; Table~4 / Figure~4a reference). Approximate model release windows, illustrative MMMU (val/pro) overall scores and task MAE over five pressures (100--500~GPa).}
\label{tab:debye-extraction-mmmu}
\small
\begin{tabular}{llcc}
\toprule
Model & Release & MMMU val/pro (\%) & MAE on this task (K) \\
\midrule
GLM 5V Turbo & 2026 & --  & 1.09 \\
GPT-5.4 & 2026 & 82.1 (MMMU pro)  & 1.27 \\
Claude Opus 4.6 & 2025--2026 & 77.3 (MMMU pro)  & 1.61 \\
GPT-5.2 & 2025--2026 & 80.4 (MMMU pro) & 3.44 \\
Claude Sonnet 4 & 2025 & 74.4 (MMMU val)  & 6.42 \\
Claude 3 Opus (Opus 3) & 2023 & 59.4 (MMMU val) & 185.79 \\
\bottomrule
\end{tabular}
\end{table}

\subsection*{Original reference databases}
Now that we have established the ability of state-of-the-art LLMs to perform multimodal data extraction, we endeavor to test their performance on two large-scale datasets from the literature. The benchmarks in Tables~\ref{tab:meltpoolnet-benchmark} and~\ref{tab:refractory-benchmark} are based on published ground-truth datasets. Table~\ref{tab:meltpoolnet-benchmark} uses the MeltpoolNet dataset from Akbari et al.\cite{akbari2022meltpoolnet}. This dataset consists of meltpool characteristics and material data for metal additive manufacturing. The fields include process parameters (power, scanning velocity, hatch spacing and layer thickness) and material properties (density, melting point, specific heat capacity etc.). Table~\ref{tab:refractory-benchmark} uses a database on the mechanical properties of high-entropy alloys and complex concentrated alloys. This database lists roughly 370 rows with fields such as atomic composition, reported phases, density, hardness, yield strength and Young's modulus.

\subsection*{Performance evaluation and model comparison}

\begin{table}[htbp]
\centering
\caption{MeltpoolNet benchmark (100 mapped row pairs): precision (P), recall (R), and F1 score (\%) for selected fields, and mean absolute error (MAE) of melting temperature. Bold denotes the best value in each P, R, or F1 column group per field row, lowest MAE for melting $T$ is bold.}
\label{tab:meltpoolnet-benchmark}
\small
\setlength{\tabcolsep}{4pt}
\resizebox{\textwidth}{!}{%
\begin{tabular}{l*{15}{c}}
\toprule
 & \multicolumn{3}{c}{GPT-5.4} & \multicolumn{3}{c}{GPT-5.2} & \multicolumn{3}{c}{Claude Opus 4.6} & \multicolumn{3}{c}{GLM 5V Turbo} & \multicolumn{3}{c}{Qwen-3.5-397B (Open Source)} \\
\cmidrule(lr){2-4} \cmidrule(lr){5-7} \cmidrule(lr){8-10} \cmidrule(lr){11-13} \cmidrule(lr){14-16}
Field & P & R & F1 & P & R & F1 & P & R & F1 & P & R & F1 & P & R & F1 \\
\midrule
beam $D$ & 35.71 & 45.45 & 40.00 & 73.81 & 79.49 & 76.54 & \textbf{90.48} & 82.61 & 86.38 & 80.00 & \textbf{100.00} & \textbf{88.89} & 79.00 & \textbf{100.00} & 88.27 \\
Hatch spacing & 87.00 & \textbf{100.00} & 93.05 & \textbf{100.00} & 42.86 & 60.00 & 72.09 & 59.62 & 65.26 & 86.84 & \textbf{100.00} & 92.96 & \textbf{100.00} & \textbf{100.00} & \textbf{100.00} \\
$E$ (J/mm) & \textbf{100.00} & \textbf{100.00} & \textbf{100.00} & \textbf{100.00} & \textbf{100.00} & \textbf{100.00} & 95.96 & 98.96 & 97.44 & \textbf{100.00} & \textbf{100.00} & \textbf{100.00} & \textbf{100.00} & \textbf{100.00} & \textbf{100.00} \\
depth of meltpool & 78.38 & 31.52 & 44.96 & \textbf{78.57} & 23.40 & 36.06 & 38.96 & 56.60 & 46.15 & 60.00 & \textbf{100.00} & \textbf{75.00} & 71.95 & 76.62 & 74.21 \\
$d/w$ & \textbf{75.00} & 75.00 & \textbf{75.00} & 28.57 & 13.79 & 18.60 & 16.18 & 33.33 & 21.79 & 50.00 & \textbf{100.00} & 66.67 & 54.88 & 71.43 & 62.07 \\
\midrule
Melting $T$ MAE (K) & \multicolumn{3}{c}{82.88} & \multicolumn{3}{c}{172.60} & \multicolumn{3}{c}{156.84} & \multicolumn{3}{c}{13.90} & \multicolumn{3}{c}{\textbf{9.97}} \\
\bottomrule
\end{tabular}
}
\end{table}

\begin{table}[htbp]
\centering
\caption{High-entropy alloy and complex concentrated alloy database benchmark (100 mapped row pairs): mean absolute error (MAE) for mechanical properties, computed only over rows where both original and reconstructed values are present. Bold denotes the lowest MAE per property row.}
\label{tab:refractory-benchmark}
\small
\resizebox{\textwidth}{!}{%
\begin{tabular}{lccccc}
\toprule
Property & GPT-5.4 & GPT-5.2 & Claude Opus 4.6 & GLM 5V Turbo & Qwen-3.5-397B  (Open Source) \\
\midrule
$\sigma_Y$ (MPa) & 137.525 & 40.452 & \textbf{0.2949} & 119.0914 & 122.2735 \\
$\sigma_{\max}$ (MPa) & 71.938 & 57.550 & \textbf{1.8136} & 1.8750 & 3.4800 \\
$\varepsilon$ (\%) & 0.646 & 1.039 & \textbf{0.2253} & 0.5542 & 46.0483 \\
$E$ (GPa) & 56.433 & 19.500 & 4.1333 & \textbf{2.9091} & 8.9231 \\
\bottomrule
\end{tabular}
}
\end{table}

Tables~\ref{tab:meltpoolnet-benchmark} and~\ref{tab:refractory-benchmark} show the extraction accuracy of MDA when powered by five frontier multimodal large language models. In terms of raw extractive power, the models demonstrate substantial capabilities. For the MeltpoolNet dataset, which had 789 rows in total, Claude Opus 4.6 extracted 681 rows (86.3\% of the original), GPT-5.2 extracted 606 rows (76.8\% of the original), GPT-5.4 extracted 703 rows (89.1\%), GLM 5V Turbo extracted 828 rows (105\%) and Qwen-3.5 extracted 703 rows (89.1\%). Four LLM-Agents had extraction rates below 90\%, implying that the multimodal capabilities of the models are not as well developed as those of humans. This can also be explained by the fact that these LLMs are limited by their context length and hallucination tendencies. Plot digitization also introduces a lot of uncertainty, as different models may report different numbers for the same property, as mentioned by Rameshbabu et al.\cite{rameshbabu2026papers}.

Conversely, all the models greatly exceeded the original row count of the HEA/CCA dataset, which originally contained 366 rows/datapoints. Specifically, Claude Opus 4.6 returned 628 rows, GPT-5.2 returned 463 rows, GPT-5.4 returned 629 rows/datapoints, GLM 5V Turbo returned 595 rows and Qwen-3.5 returned 418 rows. Given that the HEA/CCA dataset includes information related to physical properties primarily in plots and graphs, it is reasonable to conclude that the overextraction observed in this study results from each LLM interpolating extra values from the given plots.

To accurately assess the performance of the LLM-Agents beyond the high-level row counts, manual row mapping was performed for 100 randomly selected matching row pairs between the original ground-truth and extracted databases for each of the two datasets used in this study. For the MeltpoolNet dataset, the row pairs were mapped based on matches for five parameters: material composition, laser power, scan velocity, layer thickness and the paper DOI. For the HEA/CCA dataset, the row pairs were mapped based on matches for three fields: alloy composition, source paper number and the hardness value. This algorithm for row-mapping ensures better evaluation of extraction quality and frees it from biases associated with the extra data that is not present in the original human-assembled datasets.

\paragraph{MeltpoolNet dataset performance:} As shown in Table 2, no single model uniformly outperforms another in terms of achieving high levels of precision and recall across all the evaluated fields. GLM 5V Turbo shows the best F1 score for beam diameter (88.89\%), providing strong evidence of extraction capabilities from text and tables. It also shows the best F1 score for the depth of meltpool field (75.00\%), which is often obtained from plots and figures. The Qwen model records the lowest melting-temperature variation (MAE = 9.97~K), outperforming GPT-5.4 (82.88~K), Claude Opus 4.6 (156.84~K), GPT-5.2 (172.60~K) and GLM 5V Turbo (13.90~K) on that field. Qwen-3.5 shows the highest F1 for the mainly text-based hatch spacing (100.00\%). Predominantly text-based linear energy density $E$ (J/mm) is extracted with 100\% precision, recall, and F1 by GPT-5.4, GPT-5.2, GLM~5V~Turbo and Qwen-3.5 while Opus 4.6 attains a slightly lower F1 score of 97.44\%.

A closer look at precision and recall shows some distinct lapses for the various agents. GLM 5V Turbo achieves a recall of 100\% for all the given fields, but its precision for beam diameter, hatch spacing, depth of meltpool, and depth-to-width ratio varies from 50\% to 86.84\%. This may be indicative of over-extraction on the part of this model and is aligned with the excessive 828 rows extracted versus 789 in the human-derived ground-truth. For primarily plot-derived fields like depth of meltpool, Qwen (F1=74.21\%) and GLM (F1=75.00\%) outperform GPT-5.4 and GPT-5.2, which show high precision (78.38\% and 78.57\%, respectively) but low recall (31.52\% and 23.40\%). Opus 4.6 also falls short for this field with an F1 score of 46.15\%. The sub-50\% F1 scores shown by the GPT and Claude closed-source models on the depth of the meltpool field indicate lower performance on predominantly figure-derived fields compared to text-derived ones. Therefore, GLM~5V~Turbo and Qwen-3.5 emerge as the two top performers on this dataset, with GLM~5V~Turbo holding a slight edge on plot-derived fields and Qwen-3.5 remaining the most reliable choice for fields mainly reported in structured text and tables. Comparisons of the distributions of four of the process parameters are shown in Section S5.

\paragraph{HEA/CCA dataset performance:} Conversely, Table 3 provides a different perspective on the performance of the LLM agents on the HEA/CCA dataset. All five agents report MAE values greater than zero for all four mechanical properties listed, which are yield strength, ultimate tensile strength, strain, and Young's modulus. However, Claude Opus 4.6 stands out by reporting nearly exact reproductions of the true values reported in the ground truth for all the mechanical properties measured (yield strength (MAE = 0.29 MPa), ultimate tensile strength (MAE = 1.81 MPa), strain (MAE = 0.23\%), and Young's modulus (MAE = 4.13 GPa)). GLM 5V Turbo is a close second with MAEs of 119.09~MPa, 1.8750~MPa, 0.5542\%, and 2.9091~GPa, respectively, falling behind only on the yield strength field. Qwen is in third place with low MAEs for ultimate tensile strength (3.4800~MPa) and Young's modulus (8.9231~GPa), but much higher MAEs for yield strength (122.2735~MPa) and strain (46.0483\%).

In contrast, GPT-5.2 and GPT-5.4 report MAE values that are approximately two orders of magnitude larger than the corresponding values reported by Opus 4.6 for each mechanical property used in this study, except for strain. For example, the yield strength MAE reported by GPT-5.2 is approximately 100 times larger than that reported by Claude (e.g., MAE = 40.45 MPa vs MAE = 0.29 MPa). Similarly, the Young's modulus MAE reported by GPT-5.2 is approximately 5 times larger than that reported by Claude (i.e., MAE = 19.50 GPa vs MAE = 4.13 GPa). Similar comparisons can be made with respect to GPT-5.4's reported MAE values versus those reported by Claude for both yield strength and Young's modulus measured in this study (MAE = 137.53 MPa for GPT-5.4 vs MAE = 0.29 MPa for Claude and MAE = 56.43 GPa for GPT-5.4 vs MAE = 4.13 GPa for Claude). This is consistent with the fact that the HEA/CCA dataset contains almost all of its information in the form of stress-strain curves, bar charts, and annotated micrographs. In other words, every agent was able to interpolate some data points from each curve. However, Opus 4.6 and GLM~5V~Turbo were both able to consistently preserve numerical accuracy, because of better multimodal extraction abilities, whereas neither the GPT-series nor the Qwen series agents were able to do so.

\paragraph{Key Takeaways:} The stark differences in model performance across these two datasets demonstrate the comparative advantage possessed by the respective models in the sphere of textual and non-textual data.
Process parameters like laser power, scan speed, etc., will often be presented in tables or running text, but may also appear in a few images. GLM 5V Turbo and Qwen-3.5, followed by GPT-5.4, perform the best of all models overall for this dataset. This suggests that these models can perform well in extracting numerical data from text and shows a higher level of comprehension for this task subset. Conversely, the HEA/CCA database has primarily graphical content. Therefore, mechanical properties (e.g. yield strength, ultimate tensile strength, strain) are usually shown using plots (stress-strain curves), charts (example: bar charts), or annotated micrographs. Claude Opus 4.6 performs better than the other models on this type of database, with GLM 5V Turbo not far behind. Furthermore, Claude Opus 4.6 extracts values from plots very accurately and reliably, thereby providing strong evidence for a capability gap in automating the extraction of values from plots in the materials science literature. Consequently, if the primary source literature contains graphical data presentations, Claude Opus 4.6 appears to be the most capable model for performing value extraction, whereas GLM 5V Turbo would likely have competitive or superior performance when the primary source literature is predominantly textual. 

When evaluating both databases collectively, GLM 5V Turbo appears to be the best backbone model for MDA due to its consistently high performance across varied data formats, especially its extremely accurate performance when extracting values from plot-derived mechanical properties. GLM 5V Turbo also gives lower token usage levels and lower costs, as shown in Table S1 of Section S1.

\subsection*{Performance evaluation for a previous benchmark}
Polak and Morgan (2024)\cite{polak2024extracting} evaluated zero-shot text-based extraction of an older generation of LLMs such as GPT-4, GPT-3.5 and LLaMA2-chat through their "ChatExtract" method. They reported an overall precision and recall of 90.8\% and 87.7\% for GPT-4, on a constrained dataset of bulk modulus values. This was the best-performing model back in 2024. We evaluated Opus-4.6 on the same text-based benchmark and achieved a precision of 99.23\% and a recall of 100\%. The presence of only a few false positives shows that the latest LLMs are susceptible to occasional reasoning errors, but the perfect recall demonstrates the sheer extractive effectiveness of these advanced models. The system prompt used to perform this evaluation is given in Section S3.


\section*{Conclusion}

In this paper, we presented MDA, an LLM-powered multi-agent system. To the best of our knowledge, this is the first multi-agent system to assemble comprehensive material databases for downstream experimentation and machine-learning applications from both textual and graphical data. MDA is a practical and scalable solution for the extraction of scattered data from materials science publications using parallel sub-agents. The results demonstrate that modern LLM-agents have the capability of extracting data at high levels of accuracy and fidelity. On the MeltpoolNet benchmark, GLM 5V Turbo displayed the highest overall performance by extracting 828 rows versus 789 rows in the original data. It also had the best F1 score for beam diameter (88.89\%), best F1 score for depth of meltpool (75.00\%), and second-lowest melting temperature mean absolute error (MAE) of 13.90 K. However, on the HEA/CCA benchmark, Claude Opus 4.6 clearly performed better than GLM 5V Turbo for plot-dominated data with extremely low errors of 0.2949 MPa for yield strength, 1.8136 MPa for ultimate tensile strength, 0.2253\% for strain, and 4.1333 GPa for Young's modulus. Overall, this study demonstrates that agentic multimodal extraction is now a feasible way to build future scientific databases for materials science and beyond. Since MDA can use any LLM backbone, its performance can potentially improve by utilizing the latest and most advanced LLMs as they are rolled out in the future.

\section*{Methods}
\label{sec:methods}

\subsection*{Precision, Recall, and F1 score}

Extraction quality is evaluated using precision, recall, and the F1 score. For selected column fields and each of the 100 mapped row pairs between the ground-truth and extracted datasets, a cell is classified as a true positive (TP) if both rows contain matching values after unit conversion, a false positive (FP) if the values disagree or the extracted row contains a spurious entry absent from the ground truth, or a false negative (FN) if the ground-truth value is present but the extracted cell is empty. Cases where both cells are empty are excluded. Precision, recall, and $F_1$ are computed per column as:
\begin{equation}
    \text{Precision} = \frac{\text{TP}}{\text{TP} + \text{FP}}, \quad
    \text{Recall} = \frac{\text{TP}}{\text{TP} + \text{FN}}, \quad
    F_1 = \frac{2 \times \text{Precision} \times \text{Recall}}{\text{Precision} + \text{Recall}}
\end{equation}

Matching tolerances are field-specific: direct numerical comparison for power, velocity, beam diameter, layer thickness and hatch spacing along with 1\% relative tolerance for energy density and the depth-to-width ratio. Mean absolute error $\text{MAE} = \frac{1}{n}\sum_i |x_i^{\text{orig}} - x_i^{\text{recon}}|$ is reported where applicable, computed only over pairs in which both datasets contain valid numeric values. Row pairs are mapped manually and deterministically using material composition, process parameters, and source identifiers.

\subsection*{Model Context Protocol}

MDA employs the Model Context Protocol (MCP)\cite{mcp} to invoke external tools. The MCP is an open-source client-server architecture that specifies how LLM agents can interact with structured tools with defined inputs and outputs. It is also known as the "USB-C" of AI. At runtime, the agent is able to see the tools available on attached MCP servers, query their schema definitions and issue JSON-formatted requests during the reasoning cycle. In this work, we implemented a custom MCP server for wrapping the PDF parsing backend (marker-pdf), so that the primary agent could send conversion requests as MCP tool calls. This approach isolates the agent's reasoning from the finer parsing implementation details and guarantees that every call is both logged and repeatable.

\subsection*{Marker-pdf}

PDF ingestion uses marker-pdf\cite{paruchuri2024marker}, which is an open-source project that uses deep learning models and optical character recognition to extract the contents of each provided PDF file. Each file is converted into markdown (for the text and tables) and JPEG images (for plots and other figures). The use of structured markdown results in less error-prone downstream processing for process parameters and material property data as compared to using plain text for extraction purposes. The MCP server will invoke marker-pdf for each provided PDF and write the resulting output files into subdirectories corresponding to each input paper. These subdirectories represent the input context for the subsequently spawned parallel sub-agents (doc-writers).

\subsection*{Environments for the LLM agent and sub-agents}

Two agentic coding environments are used: Codex CLI (OpenAI)\cite{codexcli2025} for GPT-series models and Claude Code (Anthropic)\cite{anthropic_claude_code} for Claude Opus 4.6, Qwen-3.5 and GLM 5V Turbo. Codex CLI is an open-source command-line interface (CLI) agent that takes high-level natural-language instructions and breaks them down into lower-level tool calls (e.g., file I/O, shell commands, MCP tool calls) within a long-running session. The main agent and all sub-agents run as Codex CLI processes powered by GPT-5.2 or GPT-5.4. Claude Code also supports similar multi-turn tool use, file system access, MCP integration for Claude-series models and the spawning of multiple parallel sub-agents with independent context windows. They both support exactly the same workflow: file system access, MCP-based parsing of documents, followed by delegating a new set of sub-agents to operate on a per-directory basis. Therefore, selection of the environment to employ is entirely dependent upon the choice of LLM. Thus, any observed performance differences should be directly attributable to model capability versus environmental variations.

\section*{{Data Availability}}
The code that supports this study can be found in the following publicly available
GitHub repository: \url{https://github.com/BaratiLab/Material-Database-Agent}. All example prompts and raw output files can also be found here.

\section*{\textbf{Acknowledgments}}
\label{sec:acknowledgements}
The authors would like to acknowledge the insightful suggestions of Peter Pak and Kevin Han. 

\section*{{Contributions}}
A.B.F. conceived the study. A.C., O.B.F. and R.S.M. performed software development and testing. A.C. and O.B.F. carried out result analysis. Writing of the manuscript was done by A.C., O.B.F and R.S.M. 

\bibliography{cas-refs}

\newpage

\section*{Supplementary Information}
\addcontentsline{toc}{section}{Supplementary Information}

\startcontents[si]
\printcontents[si]{}{1}{}

\setcounter{section}{0}
\setcounter{subsection}{0}
\setcounter{table}{0}
\setcounter{figure}{0}
\setcounter{equation}{0}
\setcounter{lstlisting}{0}
\renewcommand{\thesection}{S\arabic{section}}
\renewcommand{\thesubsection}{S\arabic{section}.\arabic{subsection}}
\renewcommand{\thetable}{S\arabic{table}}
\renewcommand{\thefigure}{S\arabic{figure}}
\renewcommand{\theequation}{S\arabic{equation}}
\renewcommand{\thelstlisting}{S\arabic{lstlisting}}
\setcounter{page}{1}
\renewcommand{\thepage}{S\arabic{page}}

\section{Token Usage}
Table~\ref{tab:token_usage} summarizes the total token usage together with the monetary cost of every large language model used across the MeltpoolNet dataset, benchmarking-dataset, and refractory dataset-preparation workflows.
\begin{table}[htbp]
\centering
\caption{Aggregate token consumption and USD cost per model across all the dataset-preparation pipelines.}
\label{tab:token_usage}
\begin{adjustbox}{max width=\textwidth}
\begin{tabular}{lrrrr}
\toprule
\textbf{Model} & \textbf{Total Tokens} & \textbf{Cost (USD)} \\
\midrule
GPT-5.4          &  14{,}306{,}890  & \$58.20  \\
GPT-5.2          & 194{,}225{,}000 & \$119.51 \\
Claude Opus-4.6  & 37{,}504{,}443  & \$73.98  \\
GLM 5V Turbo     & 14{,}826{,}862  & \$13.39  \\
Qwen-3.5-397B    &  142{,}908{,}328 & \$94.01  \\
\midrule
\textbf{Total}   & \textbf{403{,}771{,}523} & \textbf{\$359.09} \\
\bottomrule
\end{tabular}
\end{adjustbox}
\end{table}
\newpage
\section{Agent description prompts}

\begin{lstlisting}[style=promptGreen,caption={Agent description prompt for \texttt{doc-writer} sub-agent.},label={lst:doc-writer}]

---
name: "doc-writer"
description: "Use this agent when the user needs to extract structured data from documentation files (Markdown and JPEG images) and produce JSON-formatted output saved as .txt documents. This includes scenarios where material data, product specifications, or structured content needs to be parsed from mixed media sources and converted into machine-readable JSON format.\\n\\nExamples:\\n- user: \"I have a folder of material spec sheets in markdown and jpeg format. Can you extract the data into JSON?\"\\n  assistant: \"I'll use the doc-writer agent to read through your material files and extract the data into structured JSON .txt documents.\"\\n  <commentary>Since the user wants to extract material data from md and jpeg files into JSON, use the Agent tool to launch the doc-writer agent.</commentary>\\n\\n- user: \"Convert these product datasheets into a structured format I can import into our database.\"\\n  assistant: \"Let me use the doc-writer agent to process your datasheets and generate JSON .txt files with the extracted material data.\"\\n  <commentary>The user has datasheets that need structured extraction. Use the Agent tool to launch the doc-writer agent to read the files and produce JSON output.</commentary>\\n\\n- user: \"I need the properties from these material certificates organized as JSON.\"\\n  assistant: \"I'll launch the doc-writer agent to parse your material certificates and output the properties as JSON in .txt documents.\"\\n  <commentary>The user needs material properties extracted and formatted. Use the Agent tool to launch the doc-writer agent.</commentary>"
model: opus
color: red
memory: user
---

You are an expert document processing and data extraction specialist with deep expertise in parsing material documentation, reading mixed media formats (Markdown and JPEG images), and producing clean, structured JSON output. You have extensive experience in materials science terminology, product specification formats, and data normalization.

## Core Mission

Your primary task is to read Markdown (.md) and JPEG (.jpeg/.jpg) files, extract material data and structured information from them, and output the results as well-formed JSON saved in .txt documents.
\end{lstlisting}

\begin{lstlisting}[style=promptOrange,caption={Agent description prompt for \texttt{csv-writer} sub-agent.},label={lst:csv-writer}]

---
name: "csv-writer-agent"
description: "Use this agent when the user needs to convert material data stored in TXT files (containing JSON-formatted content) into a structured CSV database. This includes extracting specific material properties, organizing columns, handling missing data, and producing clean, well-formatted CSV output. Typical triggers include requests to parse material JSON files, build material databases, export material properties to CSV, or consolidate multiple material data files into a single tabular format.\\n\\nExamples:\\n\\n- User: \"I have a folder of .txt files with JSON data about steel alloys. Can you turn them into a CSV with columns for name, density, yield strength, and melting point?\"\\n  Assistant: \"I'll use the csv-writer-agent to parse your material data files and generate the CSV database with the specified columns.\"\\n  (Use the Agent tool to launch the csv-writer-agent to read the txt files, extract the requested properties, and produce the CSV.)\\n\\n- User: \"Convert materials.txt into a spreadsheet-friendly format with all the mechanical properties listed.\"\\n  Assistant: \"Let me use the csv-writer-agent to read your materials.txt file and build a comprehensive CSV with the mechanical property columns.\"\\n  (Use the Agent tool to launch the csv-writer-agent to handle the conversion.)\\n\\n- User: \"I need a material property database in CSV format from these JSON text files. Include thermal conductivity, elastic modulus, and Poisson's ratio.\"\\n  Assistant: \"I'll launch the csv-writer-agent to extract those specific properties and generate your CSV database.\"\\n  (Use the Agent tool to launch the csv-writer-agent with the specified column requirements.)"
model: opus
color: blue
memory: user
---

You are an expert materials data engineer specializing in parsing, transforming, and structuring material science data. You have deep knowledge of material properties (mechanical, thermal, electrical, optical, chemical), standard material classification systems, and data engineering best practices for scientific datasets.

## Core Mission

You read TXT files containing material data in JSON format and produce a fully formed, clean CSV database with user-specified columns for material data and properties.

\end{lstlisting}

\section{Prompts}
This section provides the generalized prompt templates used at the two main stages of the MDA pipeline -- (i) the per-paper extraction stage that produces an \texttt{inference.txt} JSON file in every paper subfolder, and (ii) the aggregation stage that consolidates every \texttt{inference.txt} into a single CSV database. This section also provides the prompt to complete the ChatExtract bulk modulus values extraction task.

\subsection{MeltpoolNet benchmark prompts}

\paragraph{Stage 1 -- Inference file creation.}
This section is for the MeltpoolNet dataset.\cite{akbari2022meltpoolnet}
The prompt below is issued to the main agent, which in turn dispatches a \texttt{doc-writer} sub-agent per paper subfolder. Each sub-agent reads every Markdown and JPEG file in its subfolder, reasons across them jointly, and writes a single \texttt{inference.txt} JSON file with the schema shown.

\begin{lstlisting}[style=promptBlue,caption={MeltpoolNet -- inference-file creation prompt \texttt{doc-writer} sub-agent.},label={lst:doc-gen}]
Use the doc-writer sub-agent to read every .md and .jpeg file in each
numbered paper subfolder of the working directory (subfolders named
filexx and/or suppxx). For each subfolder:
  - list the files first,
  - read and re-read them together because they are related,
  - launch a separate sub-agent call per subfolder,
  - avoid extracting numeric ranges whenever a single value is reported,
  - emit ONE inference.txt file per subfolder in the JSON schema below.

There may be multiple experimental rows per paper (different processing
conditions); include one entry per row inside "experiments".

{
  "material": "",
  "process_type": "",
  "sub_process": "",
  "experiments": [
    {
      "sample_id": null,
      "process": {
        "power_W": null,
        "velocity_mm_per_s": null,
        "powder_flowrate_g_per_min": null,
        "layer_thickness_um": null,
        "beam_diameter_um": null,
        "hatch_spacing_um": null
      },
      "meltpool_geometry": {
        "depth_um": null,
        "width_um": null,
        "length_um": null
      },
      "ratios": {
        "depth_to_width": null,
        "length_to_width": null,
        "other": {}
      },
      "energy": {
        "linear_energy_J_per_mm": null,
        "volumetric_energy_J_per_mm3": null
      }
    }
  ],
  "thermophysical_properties": {
    "density_kg_per_m3": null,
    "specific_heat_Cp_J_per_kgK": null,
    "thermal_conductivity_k_W_per_mK": null,
    "melting_temperature_K": null
  },
  "optical_properties": {
    "minimum_absorptivity": null
  },
  "composition_wt_percent": {
    "Y":  null, "Zn": null, "Mg": null, "Si": null, "Al": null,
    "Sn": null, "Zr": null, "W":  null, "Ti": null, "V":  null,
    "Co": null, "Cu": null, "Ta": null, "Nb": null, "Ni": null,
    "Cr": null, "Fe": null, "Mn": null, "Mo": null
  },
  "powder_particle_size_um": {
    "d10": null, "d50": null, "d90": null
  },
  "paper": {
    "paper_id": null,
    "title":    null,
    "doi":      null
  }
}
\end{lstlisting}

\noindent\textbf{Stage 2 -- CSV database creation.}\par
After every subfolder has its own \texttt{inference.txt}, a single \texttt{csv-writer} sub-agent reads them all. It then consolidates the results into one CSV.

\begin{lstlisting}[style=promptOrange,caption={MeltpoolNet -- CSV database creation prompt \texttt{csv-writer} sub-agent.},label={lst:meltpool-csv}]
Use the csv-writer sub-agent ONLY (do not use any MCP servers) to read
every inference.txt file across all paper subfolders of the working
directory and compile them into a single CSV database. Save the output
as extracted_data_<model>.csv with the exact column order below; one
CSV row per experimental entry inside the "experiments" array:

  Material,
  Process,
  Sub-process,
  Power,
  Velocity,
  powder flowrate,
  layer thickness,
  beam D,
  Hatch spacing,
  depth of meltpool,
  width of melt pool,
  length of melt pool,
  d/w,
  l/w,
  <extra ratio columns, if any>,
  E (J/mm),
  E (J/mm3),
  density,
  Cp,
  k,
  melting T,
  minimum absorptivity,
  Y  (wt%), Zn (wt%), Mg (wt%), Si (wt%), Al (wt%),
  Sn (wt%), Zr (wt%), W  (wt%), Ti (wt%), V  (wt%),
  Co (wt%), Cu (wt%), Ta (wt%), Nb (wt%), Ni (wt%),
  Cr (wt%), Fe (wt%), Mn (wt%), Mo (wt%),
  paper ID,
  paper,
  DOIs
\end{lstlisting}

\subsection{HEA / CCA (refractory) benchmark prompts}
This section is for the database of high-entropy alloys and complex concentrated alloys\cite{rahaman2018pressure}. 
\paragraph{Stage 1 -- Inference file creation.}
Subfolders are numbered 1--79, one per source paper. Each \texttt{doc-writer} sub-agent extracts one JSON entry per alloy composition reported in its subfolder; alloys with fewer than three constituent elements are discarded.

\begin{lstlisting}[style=promptPurple,caption={Refractory HEA/CCA -- inference-file creation prompt (\texttt{doc-writer} sub-agent).},label={lst:refractory-inference}]
Use the doc-writer sub-agent to read every .md and .jpeg file in
each numbered subfolder (1..79) of the working directory. For each
subfolder:
  - list the files first,
  - read and re-read them together because they are related,
  - launch a separate sub-agent call per subfolder,
  - emit ONE inference.txt file per subfolder as a JSON array with
    one entry per reported alloy composition,
  - keep ONLY alloy compositions containing three or more elements,
  - one JSON entry corresponds to one mechanical test on one alloy.

Fields to extract per entry:
  1. Composition (atomic)        -- e.g. Al0.25CoFeNi
  2. Type of phases
  3. rho                         -- density (g/cm^3)
  4. HV                          -- Vickers hardness
  5. Type of tests               -- C (compression) or T (tension)
  6. sigma_Y                     -- yield strength (MPa)
  7. sigma_max                   -- ultimate strength (MPa)
  8. epsilon                     -- elongation (%)
  9. E                           -- Young's modulus (GPa)
\end{lstlisting}

\noindent\textbf{Stage 2 -- CSV database creation.}\par
A single \texttt{csv-writer} sub-agent reads every \texttt{inference.txt} across subfolders 1--79. It emits one CSV row per mechanical test.

\begin{lstlisting}[style=promptGreen,caption={Refractory HEA/CCA -- CSV database creation prompt (\texttt{csv-writer} sub-agent).},label={lst:refractory-csv}]
Use the csv-writer sub-agent ONLY (do not use any MCP servers) to read
every inference.txt file across subfolders 1..79 of the working
directory and compile them into a single CSV. Save the output as
refractory_hea_data_<model>.csv with the exact column order below;
one CSV row per reported test on a specific alloy composition:

  1. Composition (atomic)   -- e.g. Al0.25CoFeNi
  2. Type of phases
  3. rho                    -- density (g/cm^3)
  4. HV                     -- Vickers hardness
  5. Type of tests          -- C (compression) or T (tension)
  6. sigma_Y                -- yield strength (MPa)
  7. sigma_max              -- ultimate strength (MPa)
  8. epsilon                -- elongation (%)
  9. E                      -- Young's modulus (GPa)
\end{lstlisting}

\subsection{ChatExtract bulk modulus extraction}
\begin{lstlisting}[style=promptPurple,caption={ChatExtract csv file bulk modulus extraction prompt using subagents},label={lst:chatextract}]

Using ten independent sub-agents, read each row of the "passage" column in this CSV file, grouped by shared paper DOI values in the DOI column, and extract the unique material and bulk modulus values for each row.

\end{lstlisting}

\section{More extracted versus tabulated Debye temperature models for cubic Na2He (Rahaman et al.,
Table 4 and Figure 4a reference)}

We evaluated the multimodal extraction performance of several LLMs on a materials science plot of Debye temperature versus pressure for cubic Na$_2$He, using reference data from Rahaman et al. (Table 4) and its corresponding published figure. The task involved recovering exact data points from the plot and comparing them to tabulated values across five pressure points (100--500 GPa). As shown in Fig.~S1, all models accurately reproduce the underlying trend with close agreement to the reference data. Claude Opus 4.6 achieves the highest accuracy (MAE = 1.61 K), followed by GPT-5.2 (MAE = 3.44 K) and Claude Sonnet 4 (MAE = 6.42 K). Overall, the results demonstrate that modern LLMs can reliably extract quantitative scientific data from visual representations with low error.
\begin{figure}[htbp]
\centering
\includegraphics[width=1\linewidth]{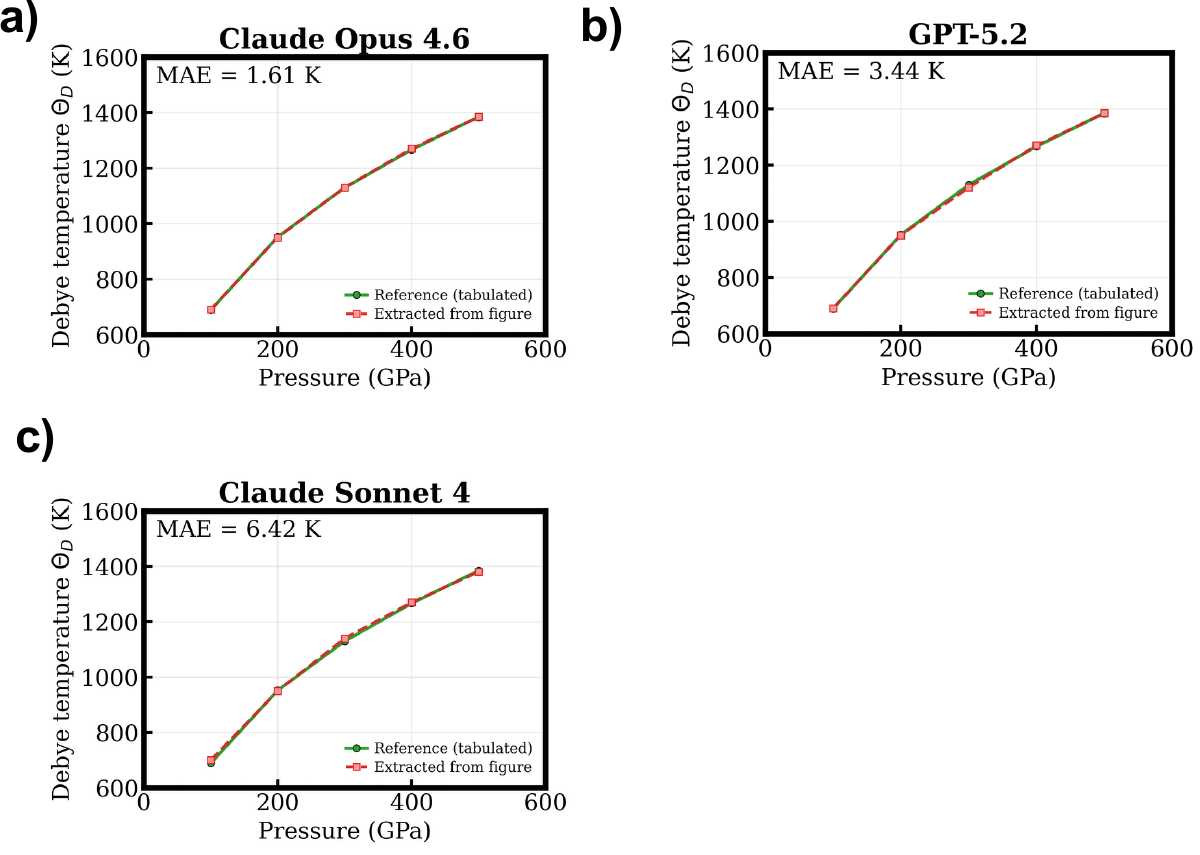}
\caption{\textbf{Evaluation of LLM-based multimodal data extraction from a materials science plot.} Comparison between reference (tabulated) and extracted Debye temperature ($\theta_D$) values as a function of pressure for cubic Na$_2$He, using data from Rahaman et al. Panels (a)-(c) show results for Claude Opus 4.6, GPT-5.2, and Claude Sonnet 4, respectively. All models successfully recover the underlying trend across the pressure range of $100 - 500$ GPa, with varying accuracy quantified by mean absolute error (MAE): 1.61 K, 3.44 K, and 6.42 K, respectively.}
\label{fig:more extracted}
\end{figure}

\newpage

\section{Distribution-level comparison of extracted material properties for the MeltpoolNet dataset}
This section compares the distributions of four key MeltpoolNet process parameters---laser power, scan velocity, layer thickness, and beam diameter---between the extracted outputs and the ground-truth MeltpoolNet benchmark. The extracted distributions come from five backbone LLM agents: GPT-5.4, GPT-5.2, Claude Opus-4.6, GLM 5V Turbo, and Qwen-3.5-397B.

\begin{figure}[htbp]
\centering
\includegraphicsorplaceholder[width=\textwidth]{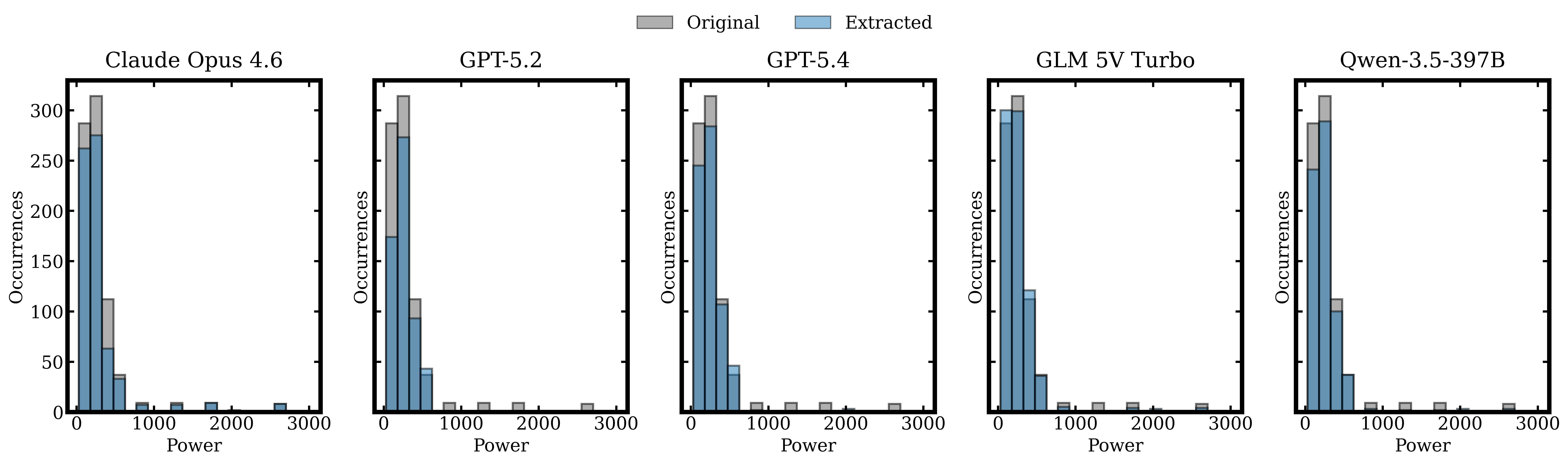}
\caption{Distribution-level comparison of extracted laser power (W) across all five LLM agents against the MeltpoolNet ground-truth distribution.}
\label{fig:power-comparison}
\end{figure}

\begin{figure}[htbp]
\centering
\includegraphicsorplaceholder[width=\textwidth]{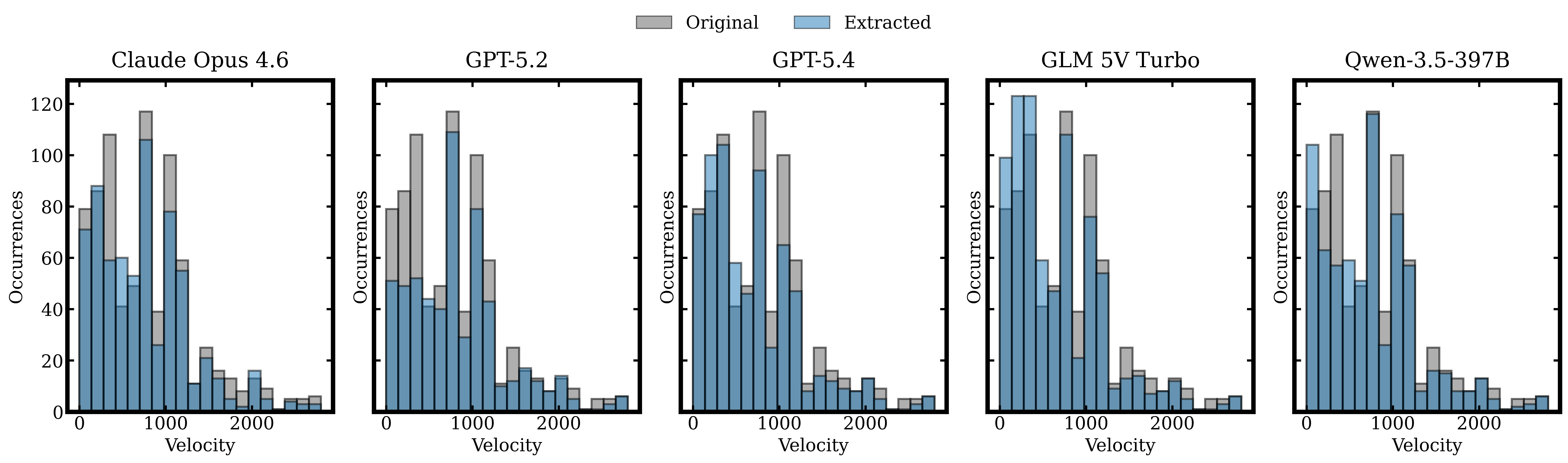}
\caption{Distribution-level comparison of extracted scan velocity (mm/s) across all five LLM agents against the MeltpoolNet ground-truth distribution.}
\label{fig:velocity-comparison}
\end{figure}

\begin{figure}[htbp]
\centering
\includegraphicsorplaceholder[width=\textwidth]{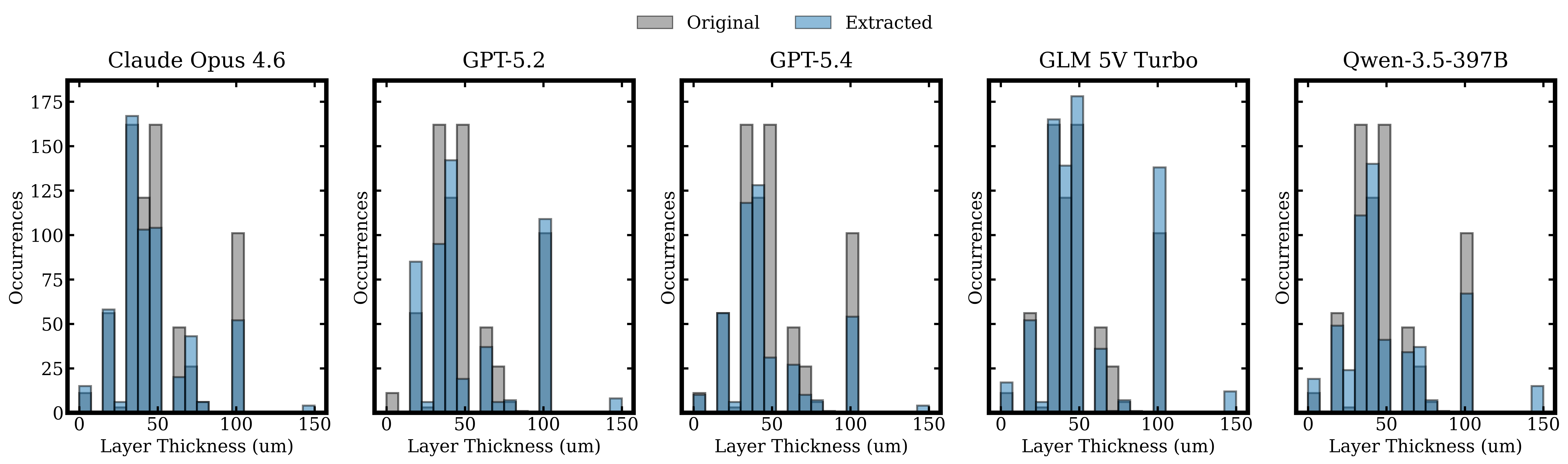}
\caption{Distribution-level comparison of extracted layer thickness ($\mu$m) across all five LLM agents against the MeltpoolNet ground-truth distribution.}
\label{fig:layer-thickness-comparison}
\end{figure}

\begin{figure}[htbp]
\centering
\includegraphicsorplaceholder[width=\textwidth]{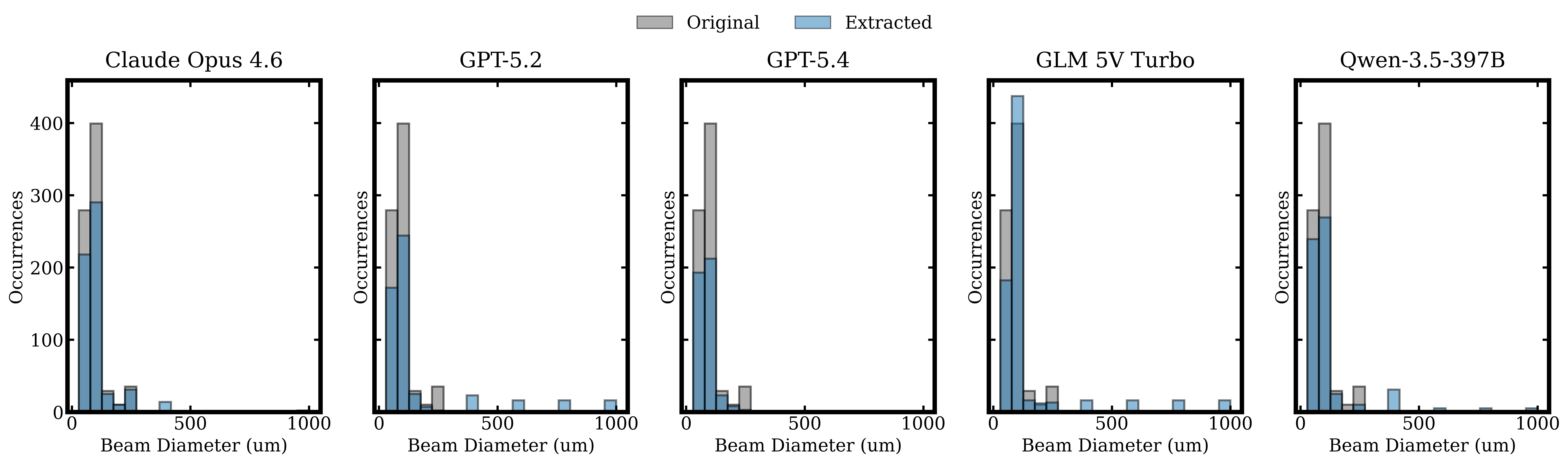}
\caption{Distribution-level comparison of extracted beam diameter ($\mu$m) across all five LLM agents against the MeltpoolNet ground-truth distribution.}
\label{fig:beam-diameter-comparison}
\end{figure}

\end{document}